\pdfoutput=1
\documentclass{article}
\usepackage{microtype}
\usepackage{graphicx}
\usepackage{subfigure}
\usepackage{booktabs} 
\usepackage{color}
\usepackage{pifont}
\usepackage{tabularx}  

\usepackage[accepted]{icml2025}

\usepackage{amsmath}
\usepackage{amssymb}
\usepackage{mathtools}
\usepackage{amsthm}
\usepackage{bm}
\usepackage{adjustbox}

\usepackage{algorithm}
\usepackage{algorithmic}
\usepackage{multirow}
\usepackage{makecell}
\usepackage{caption}
\usepackage{array}
\usepackage{xcolor}

\AtBeginDocument{%
  \providecommand\BibTeX{{%
    Bib\TeX}}}

\theoremstyle{plain}

\theoremstyle{definition}

\theoremstyle{remark}

\usepackage[disable,textsize=tiny]{todonotes}
\usepackage{hyperref}
\usepackage[capitalize,noabbrev]{cleveref}

\icmltitlerunning{TranStable: Towards Robust Pixel-level Online Video Stabilization by Jointing Transformer and CNN}
\begin{document}

\twocolumn[
\icmltitle{TranStable: Towards Robust Pixel-level Online Video Stabilization by Jointing Transformer and CNN}
\icmlsetsymbol{equal}{*}

\begin{icmlauthorlist}
\icmlauthor{zhizhen li}{sch}
\icmlauthor{tianyi zhuo}{sch}
\icmlauthor{Yifei Cao}{sch}
\icmlauthor{Jizhe Yu}{sch}
\icmlauthor{Yu liu}{sch}
\end{icmlauthorlist}

\icmlaffiliation{sch}{Dalian University of Technology, Institute of software, Dalian, China}

\icmlcorrespondingauthor{Firstname1 Lastname1}{first1.last1@xxx.edu}
\icmlcorrespondingauthor{Firstname2 Lastname2}{first2.last2@www.uk}


\vskip 0.3in
]
\begin{abstract}
Video \textcolor{green}{stab}ilization often struggles with distortion and excessive cropping. This paper proposes a novel end-to-end framework, named \textcolor{green}{\textit{TranStable}}, to address these challenges, comprising a generator and a discriminator. We establish \textcolor{green}{Tran}sformer-UNet (TUNet) as the generator to utilize the \textbf{H}ierarchical \textbf{A}daptive \textbf{F}usion \textbf{M}odule (HAFM), integrating Transformer and CNN to leverage both global and local features across multiple visual cues. By modeling frame-wise relationships, it generates robust pixel-level warping maps for stable geometric transformations. Furthermore, we design the \textbf{S}tability \textbf{D}iscriminator \textbf{M}odule (SDM), which provides pixel-wise supervision for authenticity and consistency in training period, ensuring more complete field-of-view while minimizing jitter artifacts and enhancing visual fidelity. Extensive experiments on NUS, DeepStab, and Selfie benchmarks demonstrate state-of-the-art performance.
\end{abstract}
\section{Introduction}
People use the handheld photography devices(\emph{e.g.,} smartphones and digital camcorders) to capture moments and record daily life. However, distorted fragments that degrade both the visual quality and feeling of viewing experience in video are caused by irregular shaking. Many researchers regard video stabilization as an essential preliminary step in the preprocessing of images and videos.\cite{cubuk2018autoaugment,ronneberger2015u}. It also plays a pivotal role in determining the complexity, accuracy, and robustness of downstream tasks in computer vision. Video stabilization aims to eliminate the jitter artifacts generated by camera motion, thereby producing smoother and more stable visual results. During the stabilization process of video frame recovery, missing and blurred pixel-level visual cues can introduce uncertainty into the image recovery stage. When addressing such problems, it is generally necessary to adopt regularization strategies in conjunction with prior information to constrain the solution space. Introducing additional constraints contributes to improved stabilization effects but may introduce numerous irrelevant terms during the stabilization process. 
\begin{figure}[tb]
\centering         
\includegraphics[width=240pt,height=150pt]{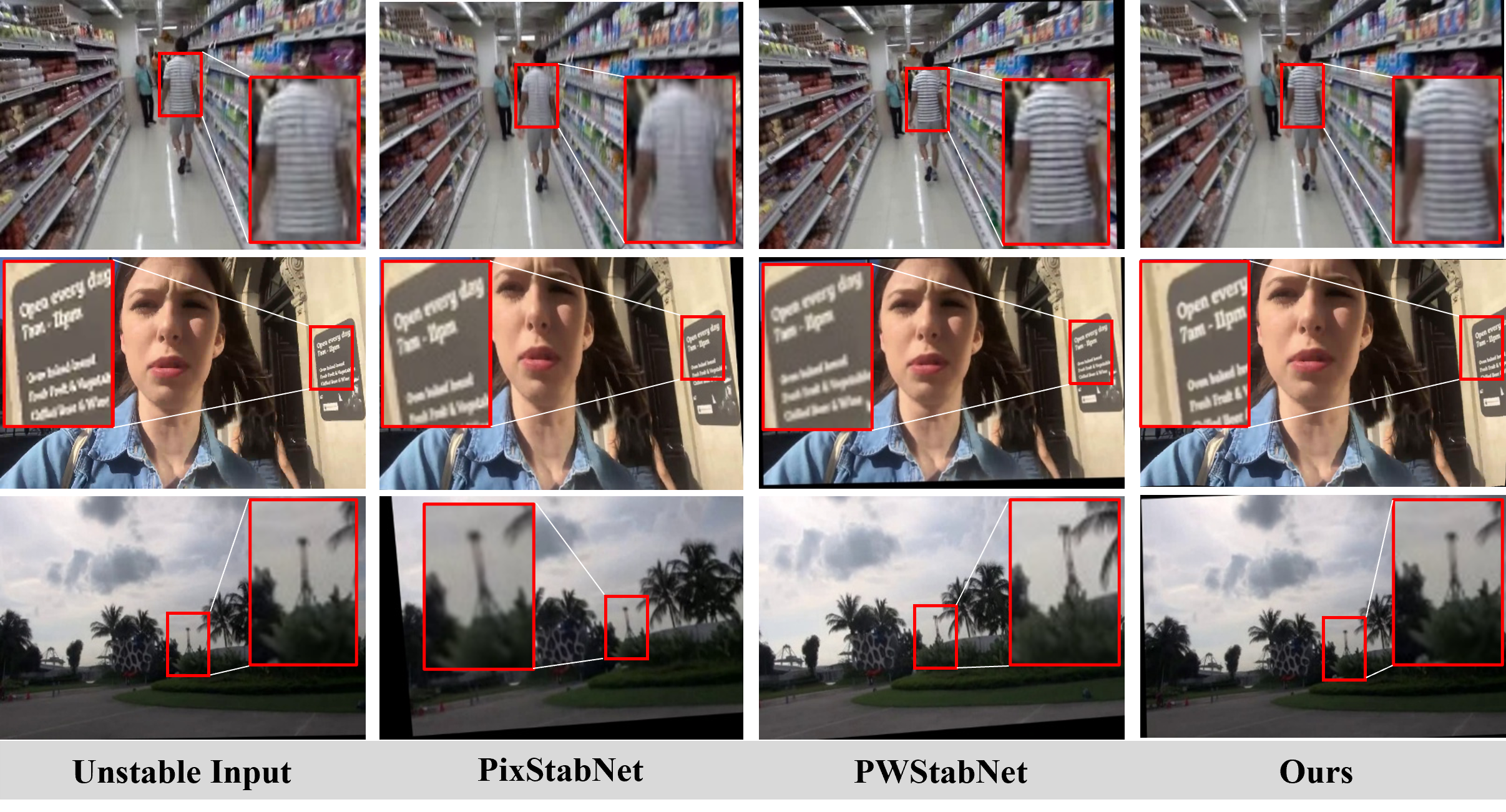}
\caption{\textbf{A comparison of our approach with state-of-the-art stabilizers}. Across various scenarios, our method exhibits a more comprehensive field of view and enhanced visual fidelity.
The locally magnified area within the red box highlights the strengths of our model.
}
\label{Fig:effect}
\end{figure}
Conventional stabilizers were employed to stabilize frames by three stages: motion estimation \cite{5031429,5974813,fan2016motion}, trajectory smoothing \cite{yu2016back}, and frame restoration. Firstly, motion-aware cues are estimated using methods like block matching \cite{8840983} or optical flow \cite{sun2018pwc}, \emph{etc.} Trajectory smoothing is applied to refine the camera trajectory. Finally, stable frames are generated using the newly computed motion trajectory\cite{NIPS2015_33ceb07b}. Specifically, it achieves effective stability by computing the homography matrix or affine transformations between frames.
\begin{figure*}[t]
\centering         
\includegraphics[height=1.5\columnwidth,width=\textwidth]{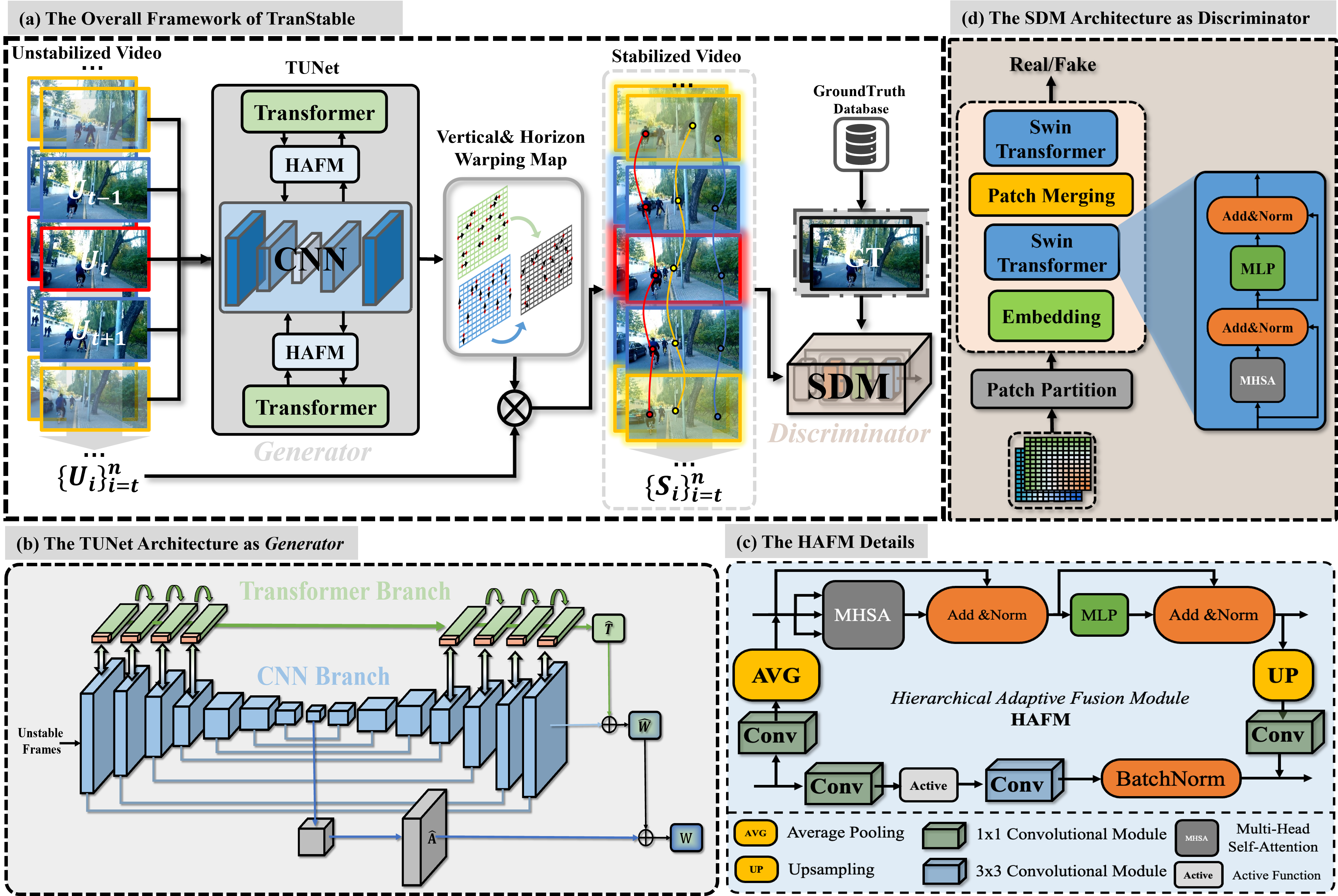}
\caption{\textbf{Overall of our approach}. (a) shows us the pipeline consists of two components: a generator and a discriminator. The generator processes consecutive video sequences to generate horizontal and vertical warping maps, which are applied to unstable frames to produce stabilized frames. The discriminator calculates the loss between the estimated frames and the ground truth frames, gradually improving the ability to generate stabilized frames. The generator utilizes a TUNet, as shown in (b) and (c), which depict the TUNet structure and highlight the HFAM component that bridges the Transformer and CNN. (d) illustrates the SDM structure, designed based on the Swin-Transformer. The seamless integration of the generator and discriminator enables the stabilization functionality, with only the generator being used during the inference stage.
}
\label{Fig:overall}
\end{figure*}
Until now, current video stabilization methods still face several challenges: local abnormal distortions, texture loss, black boarders, and excessive cropping rate. Current video stabilizers primarily rely on global motion estimation, such as motion compensation using optical flow or keypoint matching. However, these methods often struggle to handle severe jitter or local motion variations, resulting in localized distortions that cause frame deformation or structural inconsistencies. Furthermore, jitter suppression through coarse interpolation or excessive smoothing frequently blurs image details and disrupts texture consistency, leading to the loss of crucial texture information. Additionally, methods that crop frame edges to mitigate jitter often introduce blank regions, known as "black borders," and fail to effectively leverage global contextual information for proper inpainting, thereby limiting advancements in this field. Finally, many stabilizers rely on multi-frame stacking or aggressive cropping strategies, which inevitably result in excessive cropping and substantial content loss within the video frames. 

To address the aforementioned challenges, Our work draws inspiration from Generative Adversarial Networks (GANs)\cite{goodfellow2014generative}. As illustrated in \cref{Fig:overall}, the network primarily consists of a Transformer-UNet(TUNet) serving as the generator and an SDM acting as the discriminator. To the best of our knowledge, we are pioneer to apply this structure(\emph{i.e.,}GAN-like) to video stabilization, demonstrating its efficacy in this domain. For TUNet, it can integrate global scene information and local details to reconstruct stabilized frames, effectively mitigating local distortion issues. The Transformer is adept at capturing long-range dependencies within spatiotemporal sequences, making it highly effective for modeling global spatiotemporal features across multiple frames in video stabilization. This ensures accurate handling of complex jitter and errors, thereby preserving the integrity of video textures. Simultaneously, the CNN, by expanding the pixel-level receptive field, captures texture details and small-scale motion features. This capability is particularly beneficial for repairing texture loss and enhancing image sharpness, ultimately producing clearer stabilized frames. The combination of the Transformer and CNN achieves unified stabilization across both spatial and temporal domains. Additionally, the discriminator, utilizing an adversarial learning mechanism, guides the generator to incrementally enhance the realism and consistency of the generated stabilized frames. This generative approach enables the model to effectively learn latent representations of redundant black borders that require removal and to leverage contextual information for natural filling of missing regions. Although the end-to-end framework is relatively simple in structure, it operates at the pixel level. Leveraging the hierarchical adaptive fusion module, we effectively combine the strengths of the Transformer and CNN dual-branch architectures, providing a straightforward yet efficient solution to address the current bottlenecks in this field. In summary, our main contributions are as follows:
\begin{itemize}
    \item We propose a concise and efficient novel framework, termed TranStable, which achieves accurate video stabilization through the collaboration of the generator and discriminator.
    \item We introduce TUNet as generator, a model designed to generate warping maps that seamlessly integrate global context with local features. These warping maps, enriched with comprehensive semantic details from the unstable frame, enable more robust and stable geometric transformations, effectively guiding the reconstruction process in video stabilization. The core of TUNet is hierarchical adaptive fusion module (HAFM), an innovative component that combines dual branches, thereby significantly enhancing the quality of the warping maps.
    \item We propose the stability discriminator module(SDM), a discriminator built on ViT, which utilizes a pixel-wise rendering approach and a high-quality image reconstruction strategy. This component is crucial for reducing black borders and achieving a complete Field of View (FOV). 
    \item We achieve superior performance compared to representative video stabilizers across multiple public datasets.
\end{itemize}
\section{Related Work}
\subsection{Traditional Video Stabilization}
Video stabilizers estimate motion through camera trajectory, using them to reconstruct stabilized video in past few years. Selection of motion model significantly influences effectiveness, with research broadly classified into 2D and 3D methods.

\textbf{2D Video Stabilizer} typically leverage feature points or regions tracking across image pairs. \cite{zhang2015simultaneous} introduced a stabilizer to enhance stability by optimizing both low- and high-level constraints of camera path optimization. \cite{liu2014SteadyFlow} proposed SteadyFlow, achieving spatial consistency by using smooth pixel contours instead of trajectories to address varying motion. \cite{liu2016meshflow} proposed MeshFlow, an efficient online stabilizer representing sparse motion fields with grid vertices. \cite{wu2021novel} further refined motion estimation using RANSAC.

\textbf{3D Video Stabilizer} focus on mitigating camera shake by reconstructing the 3D scene and recovering camera pose. \cite{liu2009content} utilized Structure from Motion (SfM) to recover camera pose and static scene points, determining the desired camera path and subsequently calculating spatial transformations between neighboring frames. Building upon this, \cite{Zhou_2013_CVPR} introduced content preserving warping (CPW), incorporating markov random fields for segmentation and employing homography estimation to obtain warping maps.
\subsection{Video Stabilization based on Deep Learning}
Deep learning-based video stabilization trains models focusing on motion estimation, interpolation, and pixel-based methods to overcome limitations of traditional techniques like cropping artifacts and distortions.\cite{yu2019robust,smith2009light,zhao2023fast}.

The following approaches explore full-frame generation and video fidelity. \cite{choi2020deep} introduced deep iterative frame interpolation with full-frame. \cite{peng20243d} proposed a framework that integrates 3D multi-frame fusion through volume rendering. \cite{liu2009content} introduced content-preserving warping with 3D pathway. \cite{peng20243d} emphasizes the importance of 3D geometric constraints in preserving structural integrity. \cite{wang2018deep} proposed a deep online video stabilization method based on multi-grid warping transformation learning. 

\cite{wang2018deep} designed StabNet, a siamese network with an encoder and multi-grid regressor for predicting frame distortions. \cite{xu2022dut} introduced DUT, an unsupervised video stabilization method using keypoints and mesh for trajectory estimation and smoothing. \cite{choi2019difrint} developed DIFRINT, which iteratively generates intermediate frames using bidirectional optical flow, producing stable videos without cropping. \cite{yu2020learning} proposed directly generating distortion fields from optical flow. Our work draws inspiration from PWStableNet \cite{8951447}, a dual-branch network for warping map prediction. \cite{chen2021pixstabnet} presented PixStabNet, a fast online stabilizer addressing PWStableNet's multi-frame delay.

Our pipeline integrates Transformer and CNN branches via HAFM to extract and correlate visual cues that incorporate both global and local features, facilitating more accurate transformation estimation. Furthermore, the introduced SDM progressively builds correlations between the estimated frames and the groundtruth frames, steering the them towards achieving full-frame stabilization.

\section{The Proposed Method}
\subsection{The Overview Pipeline}
As shown in \cref{Fig:overall}(a), video frames are comprised of stable and unstable sequences, denoted by $\left\lbrace S_i \right\rbrace _{i=1}^n$ and $\left\lbrace U_i \right\rbrace _{i=1}^n$ respectively, where $n$ represents the total number of frames and $S, U \in \mathbb{R}^{H \times W \times C}$. TUNet takes the unstable adjacent frames $U_i$ and $U_{i+1}$ as the center of the unstable frame sequence, i.e., the sequences $\{U_{t-\Delta t}$, $U_{t+\Delta t}\}$ and $\{U_{t-\Delta t+1}$, $U_{t+\Delta t+1}\}$. The two adjacent warping maps $W_t$ and $W_{t+1}$ are generated by TUNet. To integrate dual-branches (Transformer and CNN) to learn the unstable-to-stable relationship, we design a hierarchical adaptive fusion module (HAFM) to bridge their gaps, enabling the generation of pixel-wise warping maps for steady frames. Then, Swin Transformer-based discriminator designed by us(\emph{i.e.,} SDM) to differentiate between warped and real stable frames. Training iterations refine the model's generative and discriminative capabilities. During inference, only the TUNet performs stabilization.
\subsection{Transformer-UNet(TUNet)}
As depicted in \cref{Fig:overall}(b) and (c), TUNet comprises a dual-branch architecture consisting of a UNet and a Transformer, connected by HAFM. The CNN branch incorporates seven 3x3 convolutional blocks and seven deconvolutional blocks, while the Transformer branch consists of eight Transformer blocks with identical structures. The Transformer branches operate independently, whereas the convolutional and deconvolutional blocks within the CNN branch facilitate the downward propagation of tensors.

\textbf{Transformer Branch}. Input are first embedded into the required dimensionality for the Transformer branch. The Transformer branch utilizes a 12-head multi-head attention mechanism, with the remaining architecture consistent with a standard Transformer encoder, comprising a multi-head attention layer, a feed-forward network layer, and residual normalization. As the feature map size decreases during downsampling in the CNN branch, becoming insufficient to support patch embedding, the HAFM is not employed for feature fusion in the intermediate layers.

\textbf{CNN Branch.}
The CNN branch adopts a UNet architecture, where the feature map resolution decreases with network depth while the channel count increases. As illustrated in \cref{Fig:overall}(b), the left side of CNN represents the encoder, and the right side represents decoder. The encoder captures contextual information from the image, while the decoder recovers details and generates a prediction at the same resolution as the input image. The downsampling process involves a 3x3 spatial convolution with a residual connection, followed by feature fusion with the Transformer branch via the HAFM. Batch normalization and ReLU activation are then applied. Finally, the intermediate result is passed through a FC layer to produce the first output, $\hat{\mathcal{A}}$, of the convolutional branch. Upsampling utilizes 4x4 deconvolution and, similarly, employs the HAFM for feature fusion with the Transformer branch. Skip connections are then used to recover spatial information potentially lost during the decoding stage. This process yields the second output, $\hat{\mathcal{W}}$. Finally, the output of the Transformer branch, $\hat{\mathcal{T}}$, is reshaped to match the warping map dimensions and summed with both $\hat{\mathcal{A}}$ and $\hat{\mathcal{W}}$ to produce the final output, $\mathcal{W}$.

\textbf{Hierarchical Adaptive Fusion Module(HAFM)}. Central to the TUNet architecture is the effective fusion of convolutional features from the CNN branch with the Transformer's embedded features. To achieve this, we designed the HAFM, which progressively integrates feature maps and patch embeddings. This integration leverages convolutional modules along each branch, while an asymmetric sampling stride maintains consistent embedding sizes.

\subsection{Stability Discriminator Module(SDM)}
Discriminator Module (SDM)
While TUNet generates relatively stable frame sequences after the first 20 training iterations, local texture details and object proportions can still exhibit distortions. Therefore, the SDM is designed for pixel-wise refinement and high-quality image reconstruction.  \cref{Fig:overall}(d) illustrates the structure of this simple yet effective discriminator. Inspired by SWCGAN\cite{9829280}, SDM employs Swin-Transformer v2\cite{liu2022swin} for feature extraction. We formalize scaled cosine attention as follows:
\begin{equation}
\label{eq:attention}
\textit{Attention}(\mathcal{Q}, \mathcal{K}, \mathcal{V}) = \textit{softmax}(\textit{Sim}(\mathcal{Q}, \mathcal{K})) \cdot \mathcal{V},
\end{equation}
where \(\textit{Sim}(\mathcal{Q}, \mathcal{K})\) is the scaled cosine similarity calculation function as follows:
\begin{equation}
\label{eq:sim}
\textit{Sim}(\mathcal{Q}, \mathcal{K}) = \frac{\textit{{cosine}}(\mathcal{Q} \cdot \mathcal{K})}{\tau} + \mathcal{B},
\end{equation}
where $\textit{cosine}(\mathcal{Q} \cdot \mathcal{K})$ computes the cosine similarity between the query and the key, $\tau$ is a learnable scalar that adjusts the magnitude of the similarity, and $\mathcal{B} \in \mathbb{R}^{d^2 \times d^2}$ represents the relative position bias term for each head.
\begin{algorithm}[t]
\caption{Training Stage}
\label{alg:1}
\begin{algorithmic}
    \STATE {\bfseries Input:} $S$, $U$, Feature points, $J$ \COMMENT{Source data $S$, unlabeled data $U$, and feature points}
    \STATE Initialize NetG, NetD \COMMENT{Initialize generator (NetG) and discriminator (NetD) networks}
    \FOR{$epoch = 1$ {\bfseries to} $40$} 
        \IF{$\text{use\_discriminator} = \text{true}$}
            \STATE $P_t, P_{t+1} \gets \text{NetG}(U)$ \COMMENT{Generate predicted frames $P_t$ and $P_{t+1}$ from unlabeled data $U$}
            \STATE $D_{loss} \gets \text{MSE}(NetD(P), \mathbb{P}) + \text{MSE}(NetD(S), \mathbb{S})$ 
            \COMMENT{Compute discriminator loss: MSE between predicted frames $P$ and source frames $S$}
            \STATE Update NetD \COMMENT{Update discriminator weights using backpropagation}
        \ENDIF
        \STATE Compute $\mathcal{L}_{content}$, $\mathcal{L}_{feature}$, $\mathcal{L}_{MSE}$, $\mathcal{L}_{warp}$\\
        \COMMENT{Calculate content loss, feature loss, MSE loss, and warp loss}
        \STATE $G_{loss} \gets \mathcal{L}_{content} + \mathcal{L}_{feature} + \mathcal{L}_{MSE} + \mathcal{L}_{warp}$ 
        \COMMENT{Aggregate generator loss from individual loss components}
        \STATE Update NetG \COMMENT{Update generator weights using backpropagation}
    \ENDFOR
    \STATE {\bfseries Output:} NetG, NetD \COMMENT{Return trained generator and discriminator networks}
\end{algorithmic}
\end{algorithm}
\subsection{Training}
\subsubsection{Data Pre-processing}
During training, we compute the feature point coordinates $\left\lbrace p_{x,y}^i \right\rbrace _{i=1}^{n}$ via surf\cite{bay2008speeded}, along with using RANSAC to match the feature points between two frames:$P= \{(x_i, y_i)\}_{i=1}^n$ and $P'= \{(x'_i, y'_i)\}_{i=1}^n$, $P$ represents the feature point set of the current frame, and $P'$ denotes the subsequent frame. Then, we compute the affine matrix $\left\lbrace A_{s2s}^i \right\rbrace _{i=1}^{n-1}$ between adjacent frames of a stable frame. We get the coordinate matrix $A$ as follows:
\begin{align}
A = \begin{bmatrix}
x_1 & y_1 & 1 & 0 & 0 & 0 \\
x_2 & y_2 & 1 & 0 & 0 & 0 \\
\vdots & \vdots & \vdots & \vdots & \vdots & \vdots \\
x_n & y_n & 1 & 0 & 0 & 0 \\
0 & 0 & 0 & x_1 & y_1 & 1 \\
0 & 0 & 0 & x_2 & y_2 & 1 \\
\vdots & \vdots & \vdots & \vdots & \vdots & \vdots \\
0 & 0 & 0 & x_n & y_n & 1 \\
\end{bmatrix}
\label{eq:A}
\end{align}
and construct feature point coordinate B as follows:
\begin{align}
B = \begin{bmatrix}
x'_1 & x'_2 & \cdots & x'_n & y'_1 & y'_2 & \cdots & y'_n
\end{bmatrix}^T
\label{eq:B}
\end{align}
We solve $ X $ in the following linear equation: $A \cdot X = B$, where $X = [ \alpha_{1}, \alpha_{2}, \alpha_{3}, \alpha_{4}, \alpha_{5}, \alpha_{6}]^{T}$
Next, we compute the affine matrix $\left\lbrace A_{u2s}^i \right\rbrace _{i=1}^n$ from the unstable frame to the corresponding stable frame. $A_{u2s}$ is then converted into the warping map $\lbrace \mathcal{Y}_i \rbrace _{i=1}^n$, where $A_{u2s} \in \mathbb{R}^{2 \times 3}$ and $\mathcal{Y} \in \mathbb{R}^{H \times W \times 2}$. Specifically, we normalize these coordinates and transform them into homogeneous coordinates. To ensure a minimum 15-frame separation from the video's start and end, we extract $\theta$ consecutive frames preceding and following each randomly selected training frame ($\theta$ = 15).  Here, \textit{Grid} denotes the predicted warping maps, $\mathcal{A}'$ represents the warping maps generated without the after-grid, and $U'$ and $P$ represent the frames predicted using the \textit{Grid} and $\mathcal{A}'$, respectively. Initial training focuses on the generator, with discriminator training introduced subsequently for balanced optimization.
\begin{algorithm}[tb]
    \caption{Inference Stage}
    \label{alg:stabilization}
    \begin{algorithmic}
        \STATE \textbf{Input:} Unstable frame sequence $\{U_i\}_{i=1}^n$ \COMMENT{Input sequence of unstable frames}
        \STATE Load weights: NetG \COMMENT{Load pre-trained generator weights}
        \STATE Initialize sequence: Window
        \STATE $H, W, n \gets \{U_i\}_{i=1}^n$
        \STATE Resize frames: $\{U_i\}_{i=1}^n \to \mathbb{R}^{3 \times 256 \times 256}$ \COMMENT{Resize all frames to a fixed size for processing}
        \FOR{$i = 1$ \textbf{to} $n$} 
            \IF{$i \neq 1$ \textbf{and} $i \leq n - 15$}  
                \STATE Window.pop(0)
                \STATE Window.append($U_{i+15}$) \COMMENT{Update window with future frame if possible}
            \ELSE
                \STATE Window.pop(0)
                \STATE Window.append($U_i$)
            \ENDIF
            \STATE $Grid \gets NetG(Window)$ \COMMENT{Generate warping grid using the generator network}
            \STATE $Sample \gets W(U_i, Grid)$
            \STATE $S_i \gets CropResize(H, W, Sample)$ \COMMENT{Apply cropping and resizing to stabilize the frame}
        \ENDFOR
        \STATE \textbf{Output:} Stable frame sequence $\{S_i\}_{i=1}^n$
    \end{algorithmic}
\end{algorithm}

\subsubsection{Loss Function Design}
Training details in \cref{alg:1}. Total loss of generator as below:
\begin{align}
    \mathcal{L}_{Gen} = \mathcal{L}_{con} + \alpha \cdot \mathcal{L}_{shape} + \beta \cdot \mathcal{L}_{tem},
\end{align}
where $\mathcal{L}_{con}$ represents content loss, $\mathcal{L}_{shape}$ denotes shape loss, and $\mathcal{L}_{tem}$ stands for temporal loss. The weight parameters $\alpha$ and $\beta$ are set to 1 and 8, respectively.

\textbf{Content Loss.}
Following the same setup as PixStabNet\cite{chen2021pixstabnet}, we define a content loss to minimize differences between generated stable frames and ground truth stable frames. This loss is defined as follows:
\begin{align}
    \mathcal{L}_{con} = \sum_{\substack{j=i \\ i\in [t,t+1]}} MSE(S_i - P_i) + \mathcal{L}_{VGG}(S_i , P_i),
\end{align}
where $S$ represents the ground truth, and $P$ denotes the predicted stable frame. We employ the MSE loss to penalize deviations between them. Meanwhile, we calculate the MSE loss between $S$ and $P$ in $VGG_{16}$ as below:
\begin{align}
    \mathcal{L}_{VGG} = MSE(VGG_{16}(S_i) - VGG_{16}(P_i)),
\end{align}
\begin{figure}[h]
\centering
\includegraphics[width=0.4\textwidth]{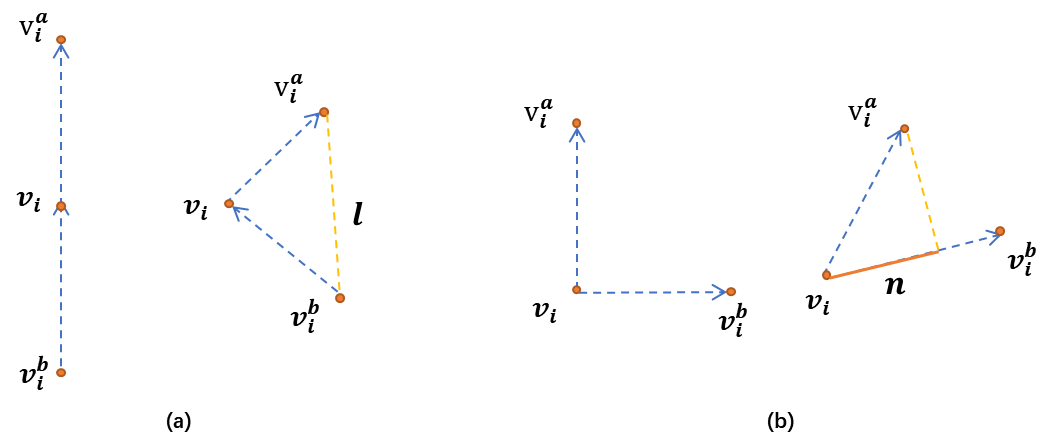} 
\caption{The relationship between two neighboring points of a point in the grid is as follows: (a) Two adjacent points are positioned on opposite sides of the center point; (b) Two adjacent points are located on the same side of the center point.}
\label{Fig:grid}
\end{figure}
\textbf{Shape Loss.}
While the predicted frames may be globally close to the ground truth, pixel-level distortions can still introduce significant artifacts in local details of the stabilized frames.  Therefore, we define a shape loss to minimize these artifacts, expressed as:
\begin{align}
    \mathcal{L}_{shape} = \mathcal{L}_{points}(\mathcal{Y}, p_i, p'_i) + \mathcal{L}_{grid}(S_i, P_i),
\end{align}
Here, $\mathcal{L}_{point}(\mathcal{Y}, S_i, P_i)$ measures the loss between corresponding feature points in consecutive frames, penalizing distortions that cause unnatural feature point movement. $\mathcal{L}_{grid}(S_i, P_i)$ represents the grid loss, defined as follows:
\begin{align}
    \mathcal{L}_{grid} = \mathcal{L}_{relative}(S_i, P_i) + \mathcal{L}_{adjacent}(S_i, P_i).
\end{align}

Stabilization model quantifies differences between sparse point correspondences in adjacent frames. It computes vector differences between four neighboring points surrounding a central point within a grid, aggregating these differences separately. $\mathcal{L}_{relative}$ quantifies changes in the relative positional relationships between these neighboring points, as illustrated in \cref{Fig:grid}(a). Formalization as below:
\begin{align}
    \mathcal{L}_{relative} = \frac{1}{n} \sum_{\substack{i\in \delta}}^n \|(v_{i}^a - v_{i}) - (v_{i} - v_{i}^b)\|_1,
\end{align}
Where $\delta$ represents the set of grid points within the frame, $n$ is the number of grid points, and $v_{i}^a$ and $v_{i}^b$ are positioned above and below, or to the left and right of $v_i$, respectively. In other words, they are the non-adjacent neighbors of $v_i$ within the grid. $\mathcal{L}_{adjacent}$ evaluates the changes between adjacent neighboring points using the dot production, as shown in \cref{Fig:grid}(b). The equation is as follows:
\begin{align}
    \mathcal{L}_{adjacent} = \frac{1}{n} \sum_{\substack{i}}^n \|(v_{i}^c - v_{i}) \cdot (v_{i}^d -v_{i})\|_1,
\end{align}
where $v_{i}^c$ and $v_{i}^d$ are adjacent neighboring points.

\textbf{Temporal Loss.}
Motion between adjacent frames is typically subtle, and the temporal smoothness of low-frequency motion is crucial. Therefore, we use a temporal loss to measure these differences.
\begin{align}
    \mathcal{L}_{temporal} = \frac{1}{HW} \|P_i - \phi (P_{i-1})\|_2^2,
\end{align}
where $\phi(\cdot)$ is a function that warps the predicted frame $P_{i-1}$ to $P_i$.
\begin{figure}[t]
\centering
\includegraphics[width=0.45\textwidth]{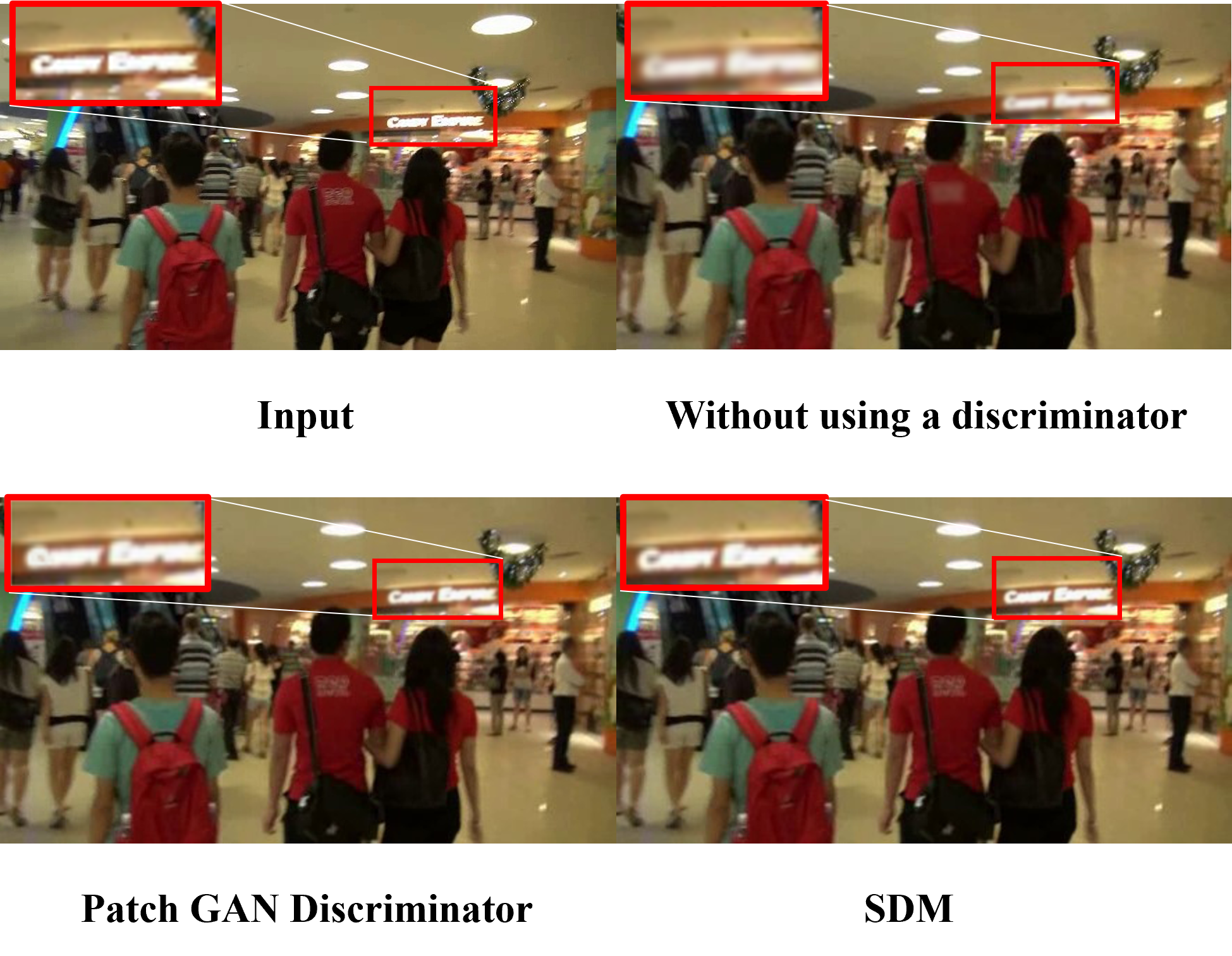}
\caption{A detailed comparison of the output frames generated using different discriminators is provided.}
\label{Fig:discriminator}
\end{figure}

\textbf{Discrimination Loss}. Meanwhile, we propose a SDM to further enhance the generation ability of stabilizing videos. Unlike previous approaches employing PatchGAN discriminators \cite{demir2018patch}, our work receives both real stable video frames as groundtruth and TUNet prediction as input. The discrimination loss as follow:
\begin{align}
    \mathcal{L}_{Dis} = \sum_{\substack{j=i \\ t < i < t+1}} (\|D(P_t) - \mathcal{P}\|_2^2 + \|D(S_t) - \mathcal{S}\|_2^2),
\end{align}
where $P_t$ represents the predicted frame, and $S_t$ denotes the ground truth stable frame. $\mathcal{P}$ and $\mathcal{S}$ are matrices filled with all -1 and all 1, respectively. 
\begin{figure*}[t]
\centering
\includegraphics[width=1.0\textwidth]{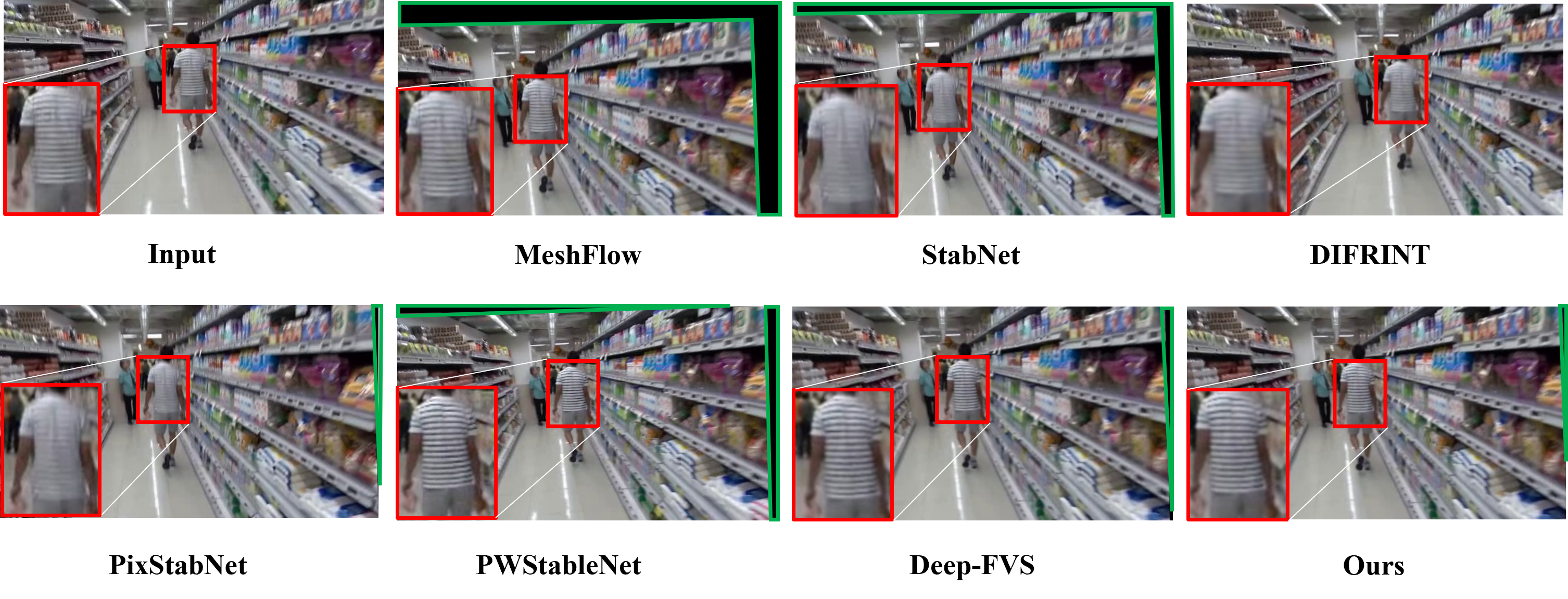}
\caption{\textbf{Quantitative analysis}. We take comparison of the output frames of certain video generated by diverse stabilizers.}
\label{Fig:framedetail}
\end{figure*}

\begin{table*}[t]
\caption{Comparison of performance on different datasets. We evaluate our method against other stabilizers using three standard metrics: Cropping Ratio (C), Distortion Value (D), and Stability Score (S). The 1st are highlighted in \textcolor{red}{red}, and the 2nd with \textcolor{blue}{blue}.}
\label{tab:performance_comparison}
\vskip 0.15in
\begin{center}
\begin{small}
\begin{sc}
\begin{tabular}{lccccccccc}
\toprule
\multicolumn{1}{c}{\multirow{2.5}{*}{\textbf{Method}}} & \multicolumn{3}{c}{NUS} & \multicolumn{3}{c}{Selfie} & \multicolumn{3}{c}{DeepStab} \\
\cmidrule(lr){2-4} \cmidrule(lr){5-7} \cmidrule(lr){8-10}
& C $\uparrow$ & D $\uparrow$ & S $\uparrow$ & C $\uparrow$ & D $\uparrow$ & S $\uparrow$ & C $\uparrow$ & D $\uparrow$ & S $\uparrow$ \\
\midrule
\multicolumn{1}{c}{Bundled \cite{liu2013bundled}} & 0.633 & 0.840 & \textcolor{blue}{0.868} & 0.682 & 0.823 & \textcolor{blue}{0.869} & 0.800 & 0.900 & \textcolor{blue}{0.850} \\
\multicolumn{1}{c}{RobustL1 \cite{grundmann2011auto}} & 0.730 & 0.817 & 0.793 & 0.751 & 0.830 & 0.807 & 0.770 & 0.870 & 0.840 \\
\multicolumn{1}{c}{Meshflow \cite{liu2016meshflow}} & 0.745 & 0.856 & 0.824 & 0.755 & 0.864 & 0.838 & - & - & - \\
\multicolumn{1}{c}{DIFRINT \cite{choi2019difrint}} & \textcolor{red}{1.000} & 0.862 & 0.807 & \textcolor{red}{1.000} & 0.868 & 0.821 & \textcolor{red}{1.000} & \textcolor{blue}{0.910} & 0.780 \\
\multicolumn{1}{c}{Yu et al. \cite{yu2019robust}} & 0.836 & 0.866 & 0.845 & 0.839 & 0.857 & 0.846 & 0.850 & 0.890 & 0.760 \\
\multicolumn{1}{c}{StabNet \cite{wang2018deep}} & 0.681 & 0.730 & 0.784 & 0.666 & 0.724 & 0.790 & - & - & - \\
\multicolumn{1}{c}{PWStableNet \cite{8951447}} & 0.854 & \textcolor{blue}{0.906} & 0.794 & 0.839 & \textcolor{blue}{0.916} & 0.803 & - & - & - \\
\multicolumn{1}{c}{Deep-FVS \cite{shi2022deep}} & 0.829 & 0.869 & \textcolor{red}{0.890} & 0.823 & 0.875 & \textcolor{red}{0.872} & - & - & - \\
\multicolumn{1}{c}{PixStabNet \cite{chen2021pixstabnet}} & 0.653 & 0.875 & 0.838 & 0.650 & 0.859 & 0.845 & - & - & - \\
\multicolumn{1}{c}{\textbf{Ours}} & \textbf{\textcolor{blue}{0.874}} & \textbf{\textcolor{red}{0.929}} & \textbf{0.866} & \textbf{\textcolor{blue}{0.865}} & \textbf{\textcolor{red}{0.942}} & \textbf{0.863} & \textbf{\textcolor{blue}{0.884}} & \textbf{\textcolor{red}{0.931}} & \textbf{\textcolor{red}{0.902}} \\
\bottomrule
\end{tabular}
\end{sc}
\end{small}
\end{center}
\vskip -0.1in
\end{table*}
\subsection{Post-processing and Inference Stage}
The top-left $(x_a, y_a)$ and bottom-right $(x_b, y_b)$ vertices of each frame are selected as key features, defining the bounding box of the frame's content. A homography matrix is computed for each frame. By analyzing the homography matrices across all stabilized frames, an initial cropping region is determined, aiming to encompass the common visible area. For each pixel in the warped (stabilized) frame sequence, its global homography is evaluated, verifying its mapping across all frames after stabilization.  The warped coordinates of each pixel are checked against the normalized range [-1, 1], representing the visible area post-stabilization. Pixels outside this range indicate regions with missing content or "holes" due to warping.  The algorithm then identifies the smallest rectangular region within [-1, 1] that is valid across all frames, ensuring no content loss after cropping. This minimum valid region defines the final cropping area, guaranteeing continuous valid content across the stabilized sequence.

The inference stage is illustrated in \cref{alg:stabilization}. Video stabilization is performed using a sliding window of length 31. The sequence is divided into three segments (the first 15 frames, the middle section, and the last 15 frames) to leverage temporal context for stabilization. Window initialization involves padding the first 15 positions with the first frame and populating the subsequent 16 positions with frames 1 through 16. The central frame within the sliding window is the current frame being processed.  The window advances by removing the leading frame and appending the frame 15 positions ahead, if available; otherwise, the final frame is appended.  Processed frames are cropped, scaled, and resized to their original dimensions before being added to the stabilized sequence.
\section{Experiments}
\subsection{Implementation Details}
During training stage, the optimizer is Adam, with parameters set to $\beta_1 = 0.5$ and $\beta_2 = 0.999$. The initial learning rate set to 0.002, which is halved every 10 epochs. After 20 epochs, the discriminator SDM is introduced to further refine the quality of the stable frames. The batch size used during training is batchsize = 8. The network is trained for 40 epochs, with the entire training process taking approximately 60 hours. In testing period, to ensure a fair comparison with previous work, the frame size is uniformly adjusted to 640×360. A Pytorch implementation of TranStable runs at 152 fps on NVidia GTX 3090Ti GPU.
\subsection{Evaluation on datasets}
To evaluate the proposed method, experiments were conducted on benchmarks, including NUS \cite{liu2013bundled}, DeepStab \cite{wang2018deep}, and Selfie \cite{choi2021self}. 
We take various comparisons with some classical and SOTA methods, including Bundled \cite{liu2013bundled}, Robust L1 \cite{grundmann2011auto}, Meshflow \cite{liu2016meshflow}, DIFRINT \cite{choi2019difrint}, Deep optical flow \cite{ilg2017flownet}, StabNet \cite{wang2018deep}, PWStableNet \cite{8951447}, Deep-FVS \cite{shi2022deep}, and PixStabNet \cite{chen2021pixstabnet}.

\cref{Fig:titcon} presents a comparative analysis of our work against other approaches using the NUS dataset, highlighting performance gap. Results reveal artifacts such as distortion and over-cropping in some methods. While subtle, differences in stabilization performance are discernible upon direct video comparison.

\subsection{Ablation Study}
We establish several variants of our method for key components. The results are shown in \cref{tab:ablation}. The TranStable baseline (Table \ref{tab:ablation}, row 1) uses a UNet to extract features from the unstable frame sequence, warps the frames using the resulting distortion map, and then predicts the refined map. Integrating features with the HAFM module, the dual-branch network improves cropping rate, distortion score, and stability score by 7.23\%, 6.43\%, and 3.79\%, respectively (Table \ref{tab:ablation}, row 2), demonstrating the effectiveness of HAFM.
We also evaluate the effectiveness of the two-layer UNet structure. In the two-layer structure, the convolutional modules in UNet pass information between layers, allowing the convolutional modules in the lower layer to integrate the processed feature information from the upper layer, thereby more accurately extracting convolutional features from the frame sequence. The model achieves performance improvements of 6.96\%, 3.81\%, and 2.78\% in cropping rate, distortion score, and stability score, respectively (see row 3 in \cref{tab:ablation}). 

Compared to the PatchGAN discriminator, SDM improves the cropping rate, distortion score, and stability score by 3.97\%, 7.40\%, and 3.20\%, respectively (see rows 4 and 5 in \cref{tab:ablation}), showing significant advantages in all three metrics. The visualization and qualitative results are shown in \cref{Fig:discriminator}, which demonstrate that SDM, compared to using no discriminator or a PatchGAN architecture, more effectively preserves original pixel and texture information, thus improving the image quality of the stabilized frames.

\begin{table}[tb]
\caption{\textbf{Ablation studies on DeepStab}. The first row in the table represents the baseline (UNet) with no additional components. Subsequent rows illustrate variations resulting from adding specific components to the baseline: HAFM only, two-layer architecture only, Patch GAN only, SDM only, HAFM + two-layer architecture, HAFM + Patch GAN, HAFM + SDM, two-layer architecture + Patch GAN, two-layer architecture + SDM, Patch GAN + SDM, HAFM + two-layer architecture + Patch GAN, and HAFM + two-layer architecture + SDM.}
\label{tab:ablation}
\vskip 0.1in
\centering
\begin{adjustbox}{width=\linewidth}  
\begin{tabular}{ccccccc}
\toprule
\multicolumn{1}{c}{\bfseries HAFM} & \multicolumn{1}{c}{\bfseries Two layer} & \multicolumn{1}{c}{\bfseries Patch GAN} & \multicolumn{1}{c}{\bfseries SDM} & \multicolumn{1}{c}{\bfseries C $\uparrow$} & \multicolumn{1}{c}{\bfseries D $\uparrow$} & \multicolumn{1}{c}{\bfseries S $\uparrow$} \\
\midrule
& & & & 0.733 & 0.762 & 0.791 \\
$\checkmark$ & & & & 0.786 & 0.811 & 0.821 \\
& $\checkmark$ & & & 0.784 & 0.791 & 0.813 \\
& & $\checkmark$ & & 0.755 & 0.783 & 0.811 \\
& & & $\checkmark$ & 0.785 & 0.841 & 0.837 \\
$\checkmark$ & $\checkmark$ & & & 0.821 & 0.833 & 0.819 \\
$\checkmark$ & & $\checkmark$ & & 0.815 & 0.839 & 0.825 \\
$\checkmark$ & & & $\checkmark$ & 0.844 & 0.851 & 0.842 \\
& $\checkmark$ & $\checkmark$ & & 0.819 & 0.826 & 0.831 \\
& $\checkmark$ & & $\checkmark$ & 0.874 & 0.896 & 0.879 \\
$\checkmark$ & $\checkmark$ & $\checkmark$ & & 0.865 & 0.907 & 0.888 \\
$\checkmark$ & $\checkmark$ & & $\checkmark$ & 0.884 & 0.931 & 0.902 \\
\bottomrule
\end{tabular}
\end{adjustbox}
\end{table}

\section{Conclusion}
This paper introduces a novel end-to-end video stabilization framework addressing distortion and cropping issues.  The generator uses a Hierarchical Adaptive Fusion Module (HAFM) integrating Transformers and CNNs to capture global and local features, generating robust warping maps.  A Stability Discriminator Module (SDM) provides pixel-level supervision for realism and consistency, minimizing jitter and maximizing field of view.  Extensive benchmark experiments (NUS, DeepStab, Selfie) demonstrate state-of-the-art results.

\textbf{Acknowledgements.} This work was supported by the Dalian Key Field Innovation
Team Support Plan (Grant: 2020RT07).

\bibliographystyle{ACM-Reference-Format}
\bibliography{ref}

\newpage
\appendix
\twocolumn
\title{Towards Generalized and Robust Pixel-level Online Video Stabilization}
\subsection{Model Architecture}
\begin{figure*}
\centering
\includegraphics[width=1.0\textwidth]{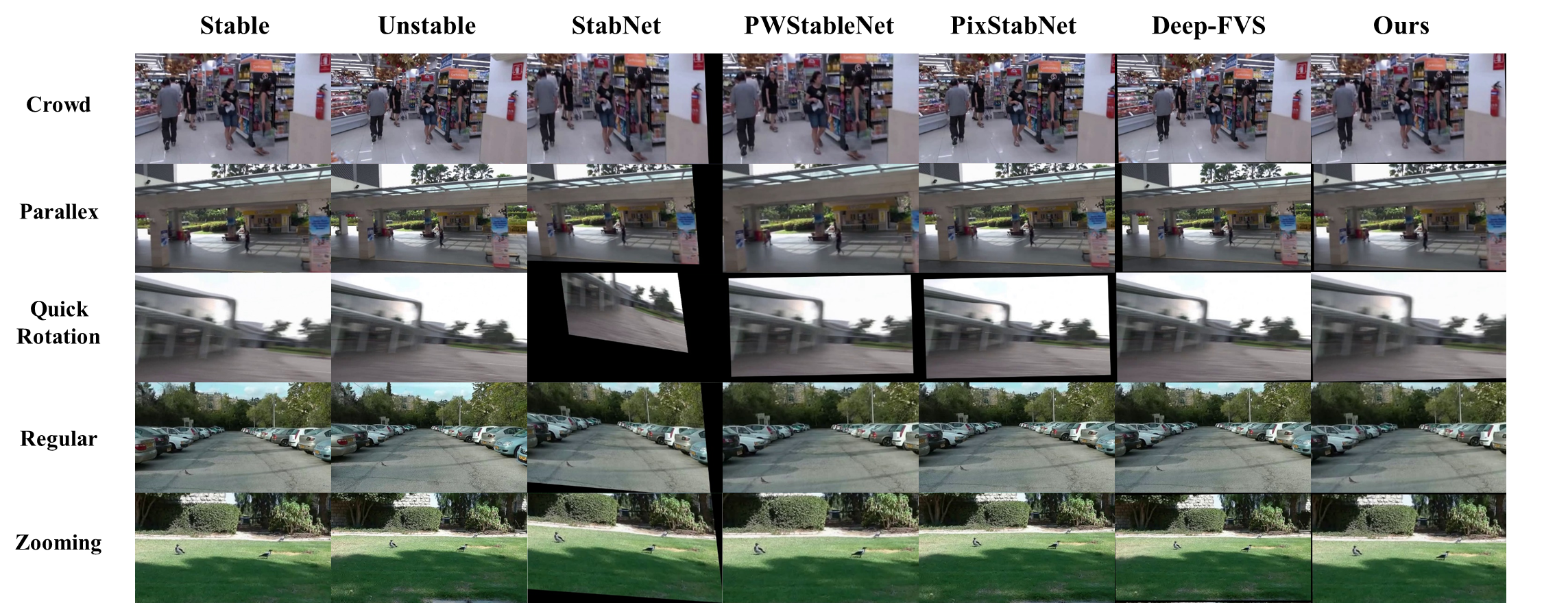}
\caption{Compared to other methods, we selected frames from various scenarios for detailed comparison.}
\label{Fig:titcon}
\end{figure*}

Detailed TUNet shown in \cref{Fig:framedetail}. Visualization example results with other stabilizers on DeepStab shown in \cref{Fig:titcon}.
\begin{table}
    \centering
    \resizebox{\linewidth}{!}{
    \begin{tabular}{ccccc}
        \toprule
        Stage 1 & Output & Conv Branch & HFAM & Trans Branch\\
        \midrule
        Init & $256 \times 256, 32$ & k = 5, p = 1, s = 2 & - & e = 768, m = 12\\
        Down1 & $128 \times 128, 64$ & k = 3, p = 1, s = 2 & $\left\{\begin{aligned}    s = 16 \rightarrow  \\        \leftarrow s = 8 \end{aligned}\right. $ & e = 768, m = 12\\
        Down2 & $64 \times 64, 64$ & k = 3, p = 1, s = 2 & $\left\{\begin{aligned}    s = 8 \rightarrow  \\        \leftarrow s = 4 \end{aligned}\right. $ & e = 768, m = 12\\
        Down3 & $32 \times 32, 128$ & k = 3, p = 1, s = 2 & $\left\{\begin{aligned}    s = 4 \rightarrow  \\        \leftarrow s = 2 \end{aligned}\right. $ & e = 768, m = 12\\
        Down4 & $16 \times 16, 256$ & k = 3, p = 1, s = 2 & $\left\{\begin{aligned}    s = 2 \rightarrow  \\        \leftarrow s = 1 \end{aligned}\right. $ & e = 768, m = 12\\
        Down5 & $8 \times 8, 256$ & k = 3, p = 1, s = 2 & - & -\\
        Down6 & $4 \times 4, 256$ & k = 3, p = 1, s = 2 & - & -\\
        Down7 & $2 \times 2, 256$ & k = 3, p = 1, s = 2 & - & -\\
        \hline
        Out($ \Hat{A_1} $) & $2 \times 3$ & \multicolumn{3}{c}{$\lbrace\begin{array}{@{}c@{}} 
        CNN: k = 2, p = 0, s = 1 \\
        MLP: 512 \rightarrow 6
        \end{array}$} \\
        \hline
        Up7 & $4 \times 4, 512$ & k = 3, p = 1, s = 2 & - & -\\
        Up6 & $8 \times 8, 512$ & k = 3, p = 1, s = 2 & - & -\\
        Up5 & $16 \times 16, 512$ & k = 3, p = 1, s = 2 & - & -\\
        Up4 & $32 \times 32, 512$ & k = 3, p = 1, s = 2 & $\left\{\begin{aligned}    s = 1
        \rightarrow  \\        \leftarrow s = 2 \end{aligned}\right. $ & e = 768, m = 12\\
        Up3 & $64 \times 64, 256$ & k = 3, p = 1, s = 2 & $\left\{\begin{aligned}    s = 2
        \rightarrow  \\        \leftarrow s = 4 \end{aligned}\right. $ & e = 768, m = 12\\
        Up2 & $128 \times 128, 128$ & k = 3, p = 1, s = 2 & $\left\{\begin{aligned}    s = 4 \rightarrow  \\        \leftarrow s = 8 \end{aligned}\right. $ & e = 768, m = 12\\
        Up1 & $256 \times 256, 64$ & k = 3, p = 1, s = 2 & $\left\{\begin{aligned}    s = 8 \rightarrow  \\        \leftarrow s = 16 \end{aligned}\right. $ & e = 768, m = 12\\
        \hline
        Out($ \Hat{W_1} $) & $256 \times 256, 2$ &  k = 3, p = 1, s = 1 & Out($ \Hat{T_1} $) & $256 \times 256, 2$\\
        
        
        \hline
        \hline
        Stage 2 & Output & Conv Branch & HFAM & Trans Branch\\
        \hline
        Down1 & $128 \times 128, 64$ & k = 3, p = 1, s = 2 & $\left\{\begin{aligned}    s = 16 \rightarrow  \\        \leftarrow s = 8 \end{aligned}\right. $  & e = 768, m = 12\\
        Down2 & $64 \times 64, 64$ & k = 3, p = 1, s = 2 & $\left\{\begin{aligned}    s = 8 \rightarrow  \\        \leftarrow s = 4 \end{aligned}\right. $  & e = 768, m = 12\\
        Down3 & $32 \times 32, 128$ & k = 3, p = 1, s = 2 & $\left\{\begin{aligned}    s = 4 \rightarrow  \\        \leftarrow s = 2 \end{aligned}\right. $  & e = 768, m = 12\\
        Down4 & $16 \times 16, 256$ & k = 3, p = 1, s = 2 & $\left\{\begin{aligned}    s = 2 \rightarrow  \\        \leftarrow s = 1 \end{aligned}\right. $  & e = 768, m = 12\\
        Down5 & $8 \times 8, 256$ & k = 3, p = 1, s = 2 & - & -\\
        Down6 & $4 \times 4, 256$ & k = 3, p = 1, s = 2 & - & -\\
        Down7 & $2 \times 2, 256$ & k = 3, p = 1, s = 2 & - & -\\
        \hline
        Out($ \Hat{A_2} $) & $2 \times 3$ & \multicolumn{3}{c}{$\lbrace\begin{array}{@{}c@{}} 
        CNN: k = 2, p = 0, s = 1 \\
        MLP: 512 \rightarrow 6
        \end{array}$} \\
        \hline
        \hline
        Up7 & $4 \times 4, 512$ & k = 3, p = 1, s = 2 & - & -\\
        Up6 & $8 \times 8, 512$ & k = 3, p = 1, s = 2 & - & -\\
        Up5 & $16 \times 16, 512$ & k = 3, p = 1, s = 2 & - & -\\
        Up4 & $32 \times 32, 512$ & k = 3, p = 1, s = 2 & $\left\{\begin{aligned}    s = 1 
        \rightarrow  \\        \leftarrow s = 2 \end{aligned}\right. $  & e = 768, m = 12\\
        Up3 & $64 \times 64, 256$ & k = 3, p = 1, s = 2 & $\left\{\begin{aligned}    s = 2 
        \rightarrow  \\        \leftarrow s = 4 \end{aligned}\right. $  & e = 768, m = 12\\
        Up2 & $128 \times 128, 128$ & k = 3, p = 1, s = 2 & $\left\{\begin{aligned}    s = 4 \rightarrow  \\        \leftarrow s = 8 \end{aligned}\right. $  & e = 768, m = 12\\
        Up1 & $256 \times 256, 64$ & k = 3, p = 1, s = 2 & $\left\{\begin{aligned}    s = 8 \rightarrow  \\        \leftarrow s = 16 \end{aligned}\right. $ & e = 768, m = 12\\
        \hline
        Out($ \Hat{W_2} $) & $256 \times 256, 2$ &  k = 3, p = 1, s = 1 & Out($ \Hat{T_2} $) & $256 \times 256, 2$\\
        \hline
        \bottomrule
    \end{tabular}
    }
    \caption{The architecture of TranStableNet includes the following components: In the convolutional branch, $\mathbf{k}$ represents the size of the convolutional kernel, $\mathbf{p}$ denotes the padding size, and $\mathbf{s}$ indicates the step size. Within the HFAM, $\mathbf{s}$ represents the step size during pooling or upsampling. In the Transformer branch, $\mathbf{e}$ signifies the embedding size, and $\mathbf{h}$ represents the number of heads in the multi-head attention mechanism.}
    \label{tbl:1}
\end{table}

\subsection{Evaluation Metrics}
The final output of the video stabilization model is a sequence of stable frames. The quality metrics for this task include the Cropping Rate Score, Distortion Score, and Stabilization Score. This paper employs these three metrics to evaluate the experimental results.
\subsubsection{Cropping rate score}
The cropping rate score represents the proportion of video images retained after cropping during the stabilization process. Due to translation, rotation, or scaling of video frames during stabilization, the output image may become smaller than the original, resulting in blank areas. The cropping rate evaluates the proportion of meaningful content retained in the frame during this process. A higher cropping ratio indicates that more significant content is preserved. For each pair of unstable and stable frames, a global homography is estimated, and the scaling factor of the homography is used to calculate the cropping ratio. Specifically, the homography matrix \(H\) is first decomposed via SVD to obtain the diagonal matrix \(\Sigma\) as follows:
\begin{equation}
\label{eq:svd}
H = U \Sigma V^T
\end{equation}
The scaling factor can be derived from the singular values, specifically by selecting the larger one as the scaling factor. In homography transformation, the singular values represent the axial scaling ratios after transformation, with the larger singular value corresponding to the greater scaling ratio. In the two-dimensional case, \(\Sigma\) is a 2×2 diagonal matrix:
\begin{equation}
\label{eq:sigma}
\Sigma = \begin{bmatrix} \sigma_1 & 0 \\ 0 & \sigma_2 \end{bmatrix}
\end{equation}
where \(\sigma_1\) and \(\sigma_2\) are the singular values of matrix \(H\), representing the scaling factors of the unit matrix. The scaling factor is the larger singular value, typically the first singular value \(\sigma_1\). The cropping rate can be expressed as:
\begin{equation}
\label{eq:cropping_rate}
\tau = \frac{1}{\sigma^2}
\end{equation}
Since a larger scaling factor \(\sigma\) results in a smaller image size after homography transformation and a larger cropped area, the cropping rate calculation must account for the proportional change of the cropped area relative to the original area. Thus, it is represented by the square of the scaling factor. To compute the average cropping rate across all frame images, assuming there are \(N\) frames with cropping rates \([\tau_1, \tau_2, ..., \tau_N]\), the final cropping rate can be calculated using the following formula:
\begin{equation}
\label{eq:avg_cropping_rate}
\tau = \frac{1}{N} \sum_{i=1}^{N} \tau_i
\end{equation}
where \(\tau\) represents the average cropping rate. For the evaluation of the stabilization task, the average cropping ratio across all frames in the video is used as the final cropping rate score.
\subsubsection{Distortion score}
The distortion score evaluates the degree of distortion between video frames, reflecting the deformation or distortion introduced by the stabilization algorithm, i.e., the difference in shape and geometric transformation between the stabilized image and the original image. This paper performs SVD decomposition on the homography matrix \(H\) and calculates the ratio of the two largest eigenvalues of the affine component. For the homography matrix \(H\), it is decomposed into the product of the affine matrix \(A\) and the non-affine matrix \(T\):
\begin{equation}
\label{eq:homography_decomp}
H = A \cdot T
\end{equation}
The affine matrix \(A\) is decomposed into eigenvalues, yielding \(\lambda_1, \lambda_2, \lambda_3\) in descending order. The distortion score is then calculated as follows:
\begin{equation}
\label{eq:distortion_score}
\delta = \frac{\lambda_1}{\lambda_2}
\end{equation}
The distortion score is calculated based on the scaling transformation of the image, reflecting the differences in scaling ratios across various directions of the image, which indicates the degree of distortion. By computing the ratio of the two largest eigenvalues of the affine matrix, an indicator reflecting the distortion level, i.e., the distortion score, is obtained. The minimum eigenvalue ratio across all frames in the video is used as the final distortion score.
\subsubsection{Stability score}
In video stabilization tasks, the stability score is used to evaluate the degree of stability, specifically the smoothness of relative motion between adjacent frames in the video. The goal of the stability score is to quantify the effectiveness of the video stabilization algorithm in reducing jitter or shaking. Typically, the trajectory paths of grid vertices extracted from the stabilized video are used for evaluation. This paper follows previous work by dividing the frame into a 4×4 grid, obtaining \(N\) vertices, and performing frequency domain analysis on these vertices over the time dimension.
The motion trajectory of each vertex in the video is represented as \(T(t)\), which denotes the trajectory of the \(i\)-th vertex at time \(t\). The trajectory path of each vertex is transformed into the frequency domain using Fourier transform, yielding its frequency domain representation. The spectrum obtained after applying the Fourier transform to the trajectory path of the \(i\)-th vertex is \(F_i(f)\), where \(f\) represents the frequency. A specific frequency range \([a, b]\) is selected, and the signal power within this range is denoted as \(P_i\):
\begin{equation}
\label{eq:signa_L_power}
P_i = \frac{\sum_{f \in [a,b]} |F_i(f)|^2}{\sum_{f} |F_i(f)|^2}
\end{equation}
Next, the percentage of the signal power within the selected frequency range \([a, b]\) relative to the entire frequency domain is calculated. The average of these percentages across all grid vertex trajectories is computed to obtain the final stability score:
\begin{equation}
\label{eq:stability_score}
S = \frac{1}{N} \sum_{i=1}^{N} P_i
\end{equation}
That is, the lowest frequency component (excluding the DC component) across the full frequency range is used as the stability score.

\end{document}